\documentclass[letterpaper, 10 pt, conference]{ieeeconf}  

\IEEEoverridecommandlockouts                              

\usepackage{graphics} 
\usepackage{epsfig} 
\usepackage{mathptmx} 
\usepackage{times} 
\usepackage{amsmath} 
\usepackage{amssymb}  
\usepackage{hyperref} 

\usepackage{subfigure}

\title{\LARGE \bf
DymSLAM:4D Dynamic Scene Reconstruction Based on Geometrical Motion Segmentation
}

\author{Chenjie Wang$^{1}$ and Bin Luo$^{1}$ and Yun Zhang and Qing Zhao$^{1}$ and Lu Yin$^{1}$ and Wei Wang$^{1}$ 
	\\and Xin Su$^{1}$ and Yajun Wang$^{1}$ and Chengyuan Li$^{1}$
\thanks{$^{1}$State Key Laboratory of Information Engineering in Surveying, Mapping and Remote Sensing, Wuhan University, Wuhan, China
        {\tt\small \{wangchenjie, luob, zhaoqing, ccyinlu, kinggreat24, xinsu.rs, yjwangisu, lichengyuan\}@whu.edu.cn}}%
}

\begin{document}

\maketitle
\thispagestyle{empty}
\pagestyle{empty}

\begin{abstract}
	Simultaneous Localization and Mapping(SLAM) is the basis for many robotic applications such as autonomous movement. Most SLAM algorithms are based on the assumption that the scene is static. However, in practice, most scenes are dynamic which usually contains moving objects, the methods based on the static assumption are not suitable. In this paper, we introduce DymSLAM, a dynamic stereo visual SLAM system being capable of reconstructing a 4D (3D + time) dynamic scene with rigid moving objects. The only input of DymSLAM is stereo video, and its output includes a dense map of the static environment, 3D model of the moving objects and the trajectories of the camera and the moving objects. We at first detect and match the interesting points between successive frames by using traditional SLAM methods. Then the interesting points belonging to different motion models (including ego-motion and motion models of rigid moving objects) are segmented by a multi-model fitting approach. Based on the interesting points belonging to the ego-motion, we are able to estimate the trajectory of the camera and reconstruct the static background. The interesting points belonging to the motion models of rigid moving objects are then used to estimate their relative motion models to the camera and reconstruct the 3D models of the objects. We then transform the relative motion to the trajectories of the moving objects in the global reference frame. Finally, we then fuse the 3D models of the moving objects into the 3D map of the environment by considering their motion trajectories to obtain a 4D (3D+time) sequence. Unlike previous attempts that have considered moving objects as outliers, and ignored them, DymSLAM obtains information about the dynamic objects. Meanwhile, DymSLAM does not rely on semantic cues or prior knowledge and is suitable for unknown rigid objects. Hence, the proposed system allows the robot to be employed for high-level tasks, such as obstacle avoidance for dynamic objects. We conducted experiments in a real-world environment where both the camera and the objects were moving in a wide range. The results confirmed that our proposed method is a state-of-the-art SLAM system for use in this dynamic environment. 
\end{abstract}

\section{INTRODUCTION}

Simultaneous Localization and Mapping (SLAM) is considered to be a fundamental capability for intelligent mobile robots. Visual SLAM, where the main sensor is a camera, has been extensively investigated in recent years. It is the core technology of several relevant applications like collisionless navigation of robots or automatic driving. However, many of the methods make the assumption of a static environment where the only motion is that of the camera~\cite{engel2017direct,forster2014svo,mur2017orb,gomez2019pl}. As a result, it is still a great challenge to work robustly in a dynamic scene where an unknown number of objects are moving independently~\cite{saputra2018visual}.

The typical approach is to identify moving regions based on semantic segmentation and to classify them as outliers with the motion of the camera relative to the static background~\cite{bescos2018dynaslam,yu2018ds}. Alternatively, some geometric or neural network based methods also remove the outliers of the camera motion model under a static scene hypothesis~\cite{jaimez2017fast,li2017rgb,tan2013robust}. These methods reduce the impact of the dynamic objects and ignore the dynamic part of the stable 3D map~\cite{scona2018staticfusion,zhang2018posefusion,zhao2018learning}. However, the dynamic object information is ignored and not acquired, which is very important in robotic applications~\cite{saputra2018visual,xu2019mid}. Therefore, handling a dynamic scene where both the camera and rigid objects are moving over a wide range in the field of SLAM should be defined as a 4D dynamic scene reconstruction problem~\cite{saputra2018visual,luiten2019track} that involves obtaining the 3D model and the trajectory of the moving object while estimating the trajectory of the camera and reconstructing a dense map of the static background, in order to achieve 4D (3D + time) scene reconstruction. To achieve this goal, some researchers estimated the motion trajectory of each object moving in the foreground around the robot and reconstructed its 3D model based on semantic cues as an initial~\cite{xu2019mid,runz2018maskfusion}. These methods based on semantic segmentation can obtain information about moving objects. However, their scope of application is limited because many moving objects are unknown and can not be semantically segmented in practical environments where this type of method is invalid. In contrast, methods based on multi-motion segmentation have also been developed~\cite{judd2018multimotion,runz2017co}, which cluster points of the same motion into a motion model parameter instance, thereby segmenting the multiple motion models corresponding to moving objects one by one in the dynamic scene. These methods estimate the trajectory of the camera and the moving objects without relying on semantic cues as initial and are more robust in real-world scenes. However, the off-the-shelf public methods based on multi-motion segmentation are unable to simultaneously output a dense map of the static background as well as dense models of the moving objects in dynamic environments where both the camera and the object are moving over a wide range~\cite{judd2018multimotion,runz2017co}. 

In this paper, we propose a stereo dynamic visual SLAM system called DymSLAM to achieve 4D dynamic scene reconstruction. DymSLAM estimates the trajectories of the camera and each rigid moving object in the global reference frame while simultaneously reconstructing their 3D point-cloud and a dense map of the static background over time. This is done by segmenting a scene into the static background and different moving objects in the foreground corresponding to multiple motion models one by one. We segment the multiple motion models of the scene by a multi-model fitting approach that clusters points that move consistently in 3D, and associate each pixel belonging to the different moving objects with a single motion model. Next, we utilize the projected masks obtained from transformation from 3D model to 2D image to improve the effect of inexact segmentation of the boundary. We then estimate the 6DOF rigid pose of each motion model and reconstruct the dense 3D point cloud of each motion model. Finally, we transform the motion of the moving objects relative to the camera and the 3D models of the moving objects into the global reference frame to obtain a 4D (3D+time) sequence. The proposed method does not rely on semantic information as initial or prior knowledge (such as~\cite{dame2013dense,salas2013slam++}), and can segment unknown moving objects based on geometrical motion segmentation. To the best of our knowledge, this is the first stereo dynamic SLAM system capable of tracking and reconstructing a 3D model of the rigid moving objects over time based on multi-motion segmentation even when both the camera and objects are undergoing a wide range movement (in our experiment is a corridor range of longer than 25m). 

The main contributions of this paper are:
\begin{enumerate}
	\renewcommand{\labelenumi}{(\theenumi)}
	\item We propose a dynamic SLAM system being capable of estimating the trajectory of the camera and the rigid moving objects in the global reference frame while simultaneously reconstructing their dense point-cloud and a dense map of the static background.
	\item Our system is based on pure geometrical motion segmentation without semantic information to segment the different motion models of the scene and obtain the accurate masks of unknown moving objects after improving the inexact segmentation of the boundary.
	\item We believe that this is the first stereo dynamic SLAM system that is capable of tracking and reconstructing 3D models of the rigid moving objects based on geometrical motion segmentation even when both the camera and objects are moving over a wide range, in order to create a 4D (3D + time) point-cloud of the dynamic scene. 
\end{enumerate}

\begin{figure*}
	\begin{center}
		\includegraphics[width=0.8\linewidth]{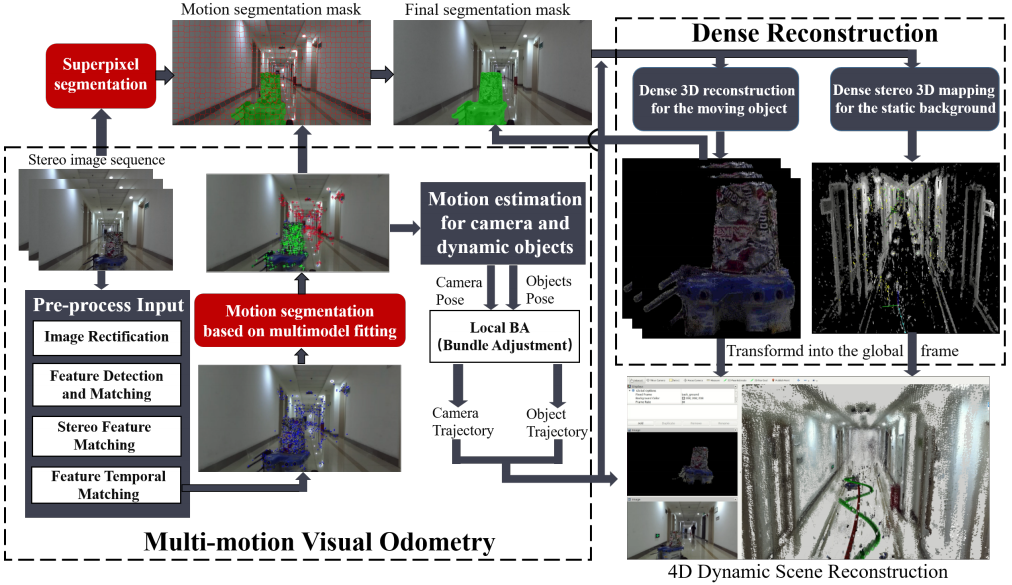}
	\end{center}
	\vspace{-0.6cm}
	\caption{Overview of the proposed dynamic SLAM system. A detailed description can be found in Section~\ref{sec:system introduction}.}
	\label{fig:fig1}
	\vspace{-0.6cm}
\end{figure*}

\section{Related Work}
The core underlying assumption behind many of the traditional visual SLAM methods is that the scene is largely static, and the only motion is that of the camera~\cite{engel2017direct,forster2014svo,mur2017orb,gomez2019pl}. To deal with dynamic scenes where the objects are always moving around the robot, some SLAM methods consider the dynamic parts as outliers to such a static model. Example of such methods are DynaSLAM~\cite{bescos2018dynaslam} and DS-SLAM~\cite{yu2018ds}. The moving objects are not further processed and their information is ignored, which restricts the autonomous movement ability of the robot in such a dynamic environment. In a dynamic scene, the robot not only needs to complete its own localization and perception of the static background, but it also needs to obtain the motion of the unknown dynamic objects and their dense models at all times. Therefore, 4D scene reconstruction should be completed to handle such dynamic scenes by obtaining the following information in the global frame: 1) the detailed 3D geometry of the moving object; 2) the motion trajectory of the object and the robot; and 3) a dense map of the static background. A number of researchers have tried to achieve these goals. In this paper, the current related work is divided into two main categories, i.e. based on semantic segmentation and multi-motion segmentation.
\subsection{Dynamic SLAM Based on Semantic Segmentation}
The method developed by the of HKUST~\cite{li2018stereo} tracks the 3D semantic objects in different motions simultaneously, instead of ignoring them, in dynamic autonomous driving scenarios. This method can handle the moving cars in a dynamic road scene, while being unable to reconstruct a dense model and address other dynamic objects except cars well. The struct2depth method developed by Google Brain~\cite{casser2019unsupervised} models the motions of individual objects precomputed by instance segmentation masks, and learns their 3D motion vector jointly with depth and ego-motion. The MaskFusion system~\cite{runz2018maskfusion} segments and assigns semantic class labels to different objects moving independently in the foreground while tracking and reconstructing them densely. This method uses a combination of instance segmentation from Mask-RCNN~\cite{he2017mask} and geometric edges to segment objects, while tracking and reconstructing the objects separately, based on ElasticFusion~\cite{whelan2016elasticfusion}. The MID-Fusion system~\cite{xu2019mid} undertakes an integrated segmentation using geometric, photometric, and semantic information. Both MaskFusion and MID-Fusion are able to segment and track multiple categories of moving objects well, and they can reconstruct a 3D representation for each rigid moving object over time. The dynamic SLAM systems based on semantic segmentation can maintain the trajectories and the 3D model of each rigid moving object. However, they rely on a semantic cue as an initial while many moving objects are unknown and not semantically segmented in practical environments where this type of method is invalid.
\subsection{Dynamic SLAM Based on Multi-motion Segmentation}
These methods segment the moving objects into different motion models in the foreground by incorporating the multi-motion segmentation method with the SLAM operation, which involves clustering points of the same motion into a motion model parameter instance. Hence, they do not rely on semantic information as initial and are more robust in a dynamic environment. MVO~\cite{judd2018multimotion} estimates the full SE(3) trajectory of both a stereo/RGB-D camera and moving objects in the dynamic scene without relying on simplifying constraints or fragile initialization. This is done by applying a multi-model fitting technique (CORAL~\cite{amayo2018geometric}) to the traditional visual odometry (VO) pipeline to segment and track each rigid object moving in the foreground. However, it does not simultaneously reconstruct a dense map of the static background and 3D models for moving objects. Stuckler and Behnke~\cite{stuckler2015efficient} proposed to segment and estimate dense rigid-body motion for the RGB-D images, but this method does not simultaneously reconstruct the objects. Finally, the Co-Fusion system~\cite{runz2017co} effectively tracks and reconstructs the 3D shape of each rigid object moving independently from the background over time while simultaneously building a map of the environment based on an RGB-D camera. It segments objects by either multi-motion segmentation or object instance segmentation~\cite{pinheiro2016learning}, and then tracks objects separately using the same approach as MaskFusion. However, it simultaneously reconstructs models of the moving objects and the static background over time only when the camera is under minor range movement or objects are about to start moving. In comparison, the proposed method maintains a dense map of the static background as well as the trajectories and 3D models of the rigid moving objects even when both the camera and the objects are moving over a wide range (in our experiment is a corridor range of longer than 25m).  
\section{System Introduction}
\label{sec:system introduction}
In this section, we describe the proposed DymSLAM method in detail, which is a new stereo visual SLAM system for 4D dynamic scene reconstruction. In a dynamic scene where both the camera and the object are under wide range movement, we solve this dynamic SLAM problem using 4D dynamic scene reconstruction, which can obtain the following information: the detailed 3D point cloud and the motion trajectories of the moving rigid objects, the ego-motion of the camera, and a dense point-cloud map of the static background. Fig.~\ref{fig:fig1} shows an overview of the proposed system. First of all, the incoming RGB stereo sequences are rectified and undistorted. Salient image features are then detected and matched across the left and right frames in each stereo pair and across temporally consecutive pairs of the stereo frames. These stereo and temporally matched feature points are then clustered into multiple motion model parameter instances by incorporating a multi-motion segmentation method based on multi-model fitting~\cite{zhang2017permutation} in the SLAM system. These motion models correspond to the motion of the camera and each moving object. Once the results of the multi-motion segmentation are stable after a few frames, each pixel in the scene is associated with a single motion model by applying the assignment problem at the superpixel level. And to compensate for inexact segmentation at the boundary, we utilize the masks projected from 3D model of the moving object. We estimate the 6DOF rigid pose of each motion model in the current frame and output the trajectories of the camera and objects in different motions after local bundle adjustment. By combining the newly estimated rigid poses, the dense 3D point cloud of each motion model is reconstructed and improved over time by fusing the points labeled as belonging to that model. Finally, we transform each object point cloud into the global reference frame with its trajectory to obtain a 4D (3D + time) point cloud of the dynamic scene.
\section{Multi-motion Visual Odometry}
This section extends our previous work~\cite{zhang2017permutation,zhao2019stereo}, which dealt with multi-motion segmentation of matched feature points~\cite{zhang2017permutation}, and also with the VO estimation of each motion model~\cite{zhao2019stereo}. We extend the traditional VO using RANSAC~\cite{mur2017orb,fischler1981random} for motion model parameter estimation to simultaneously estimate the trajectories of both the stereo camera and the moving objects. This is done by applying a multi-model fitting method to estimate the multiple motion models existing in the dynamic scene and realize VO estimation of each moving rigid target in the scene where multiple moving objects exist. This section introduces our multi-motion VO pipeline (see Fig.~\ref{fig:fig1}).
\subsection{Multi-motion Segmentation}
In this paper, we use the well-known LIBVISO2~\cite{geiger2011stereoscan} to extract and match feature points from the stereo images and use permutation preference~\cite{zhang2017permutation} of quantifying the residual error to represent the data points for the linkage clustering~\cite{magri2014t,toldo2008robust}, to segment the tracking feature points belonging to the different motions. Assume that the residual matrix calculated by the hypothetical model is R, where each column of the matrix R represents the residual value of the hypothetical model for N data points \{$r_{1,M}$,$r_{2,M}$,$\cdots$,$r_{N,M}$\}, and each row represents the residual value of each point under the M hypothetical model parameters \{$r_{N,1}$,$r_{N,2}$,$\cdots$,$r_{N,M}$\}. Next, We quantify each column of the residual matrix R separately:

\begin{equation}\label{eq1}
\check{q}_{i,j} = [\frac{r_{i,j}-r^j_{min}}{r^j_{min}-r^j_{max}}*\theta],
\end{equation}
\begin{equation}\label{eq2}
r^j_{min} = min\{r_{1,j},\cdots,r_{N,j}\}; r^j_{max} = max\{r_{1,j},\cdots,r_{N,j}\}.
\end{equation}

Where $\check{q}_{i,j}$ represents the elements in the ith and jth columns of the matrix $\check{Q}$ after the quantization of the residual matrix R. $\theta$ is the quantifying level -- in our experiments, $\forall \theta \in [100,800]$ and usually goes to 200. We use the truncation level to describe preferences by (\ref{eq3}), because the higher the quantifying value, the smaller the impact on data points.

\begin{equation}\label{eq3}
q_{i,j}=\left\{
\begin{matrix}
\check{q}_{i,j} & \check{q}_{i,j}\leq \lambda \\
0 & \check{q}_{i,j}\geq \lambda
\end{matrix}
\right.
\end{equation}

Where $\lambda$ is the quantifying length -- in our experiments, $\forall \lambda \in [1,50]$ and usually goes to 1. For the data point x, its quantifying residual preference representation is the i-th row of the obtained truncated quantifying residual matrix Q, which is $q_{i,j} = \{q_{i,1},q_{i,2},\cdots,q_{i,M}\}$.

The above is the method of using quantifying residuals to represent data points. When the hypothetical model needs to be represented, the residual matrix is first transposed. And the same operation can be performed on the transposed matrix to obtain the preference representation of the quantifying residual of the hypothetical model. Unlike other approaches that initially sample hypotheses from the scene alone, we iteratively alternate between sampling the hypotheses within the clusters and clustering the points with the permutation preference~\cite{zhang2017permutation}. After classifying the inliers belonging to the different motion models, robust estimation of each motion model parameter is then performed using RANSAC. The inliers belonging to each motion model parameters can later be used to motion estimates of the camera and the moving objects by identifying the static part of the scene.
\subsection{Multi-motion Estimation}
In this section, we describe how the inliers belonging to different motion models are converted into independent rigid motions and the trajectories of the camera and each moving rigid object~\cite{zhao2019stereo}. In continuous image sequences, we first assign an accurate and stable motion label obtained from multi-motion segmentation to each moving object. This is done by using a joint label association method based on sliding window~\cite{zhao2019stereo}. For the current frame and adjacent n key frames (in our experiment n takes 4) within the sliding window, we pass labels $l(i)$ from the feature points to the matched points. In the current frame, we take the label with the maximum coefficient:

\begin{equation}\label{eq4}
\max\limits_{l(i)}(\sum_{i=1}^n{w(i)}),
\end{equation}
where $w(i)$ represents the weights corresponding to different key frames, which decrease as the time interval increases.

Then, the motion trajectory of each label is estimated through the use of the traditional VO batch estimation technique, using only a rigid-body assumption. We track the six degrees of freedom rigid pose of each moving object with a single motion model by minimizing an energy function with a geometric iterative closest point (ICP) error between the corresponding feature points belonging to that specific model in the current frame and the 3D feature points cluster aligned with the pose in the previous frame. We use the position of the camera in the first frame as the global reference frame, and the motion trajectories of the camera and the moving objects in the global reference frame are estimated after identifying a model to represent the motion of the camera $T_C$. 
For each moving object, the 3D visual features are projected into the first frame according to the estimated pose to calculate the gravity center of the surface point set of the moving object. The result is considered to be the initial transform $T_{init}$ relating each moving object to the camera. The gravity center is adjusted and updated by considering new points due to movement over time. The motions $T_{M_tM_1}$ of each moving object with a single motion model in the global reference frame can be obtained by 

\begin{equation}\label{eq5}
T_{M_tM_1} = T_{C_tC_1}{^{ego}T_{M_tM_1}^{-1}}T_{init}^{-1},
\end{equation}

where ${^{ego}T_{M_tM_1}}$ represents the motion of the camera relative to the moving object with a single motion model. $T_{C_tC_1}$ is the motion of the camera at time instant t.

\section{Moving object mask}
\subsection{Label Assignment}
We estimate the absolute pose of each moving object at time t in the global reference frame, which is represented by the rigid transformations $T_{M_tM_1}$ after the VO pipeline. In addition, the motion of the camera with respect to the global reference frame at time t is described by rigid transformations $T_{C_tC_1}$. In this section, each point of the subsequent frame is assigned to a single label by associating it with the motion of one of the rigid models. In order to complete efficient motion segmentation pixel-by-pixel, we apply the labeling algorithm at the superpixel level based on  incorporating superpixel segmentation (simple linear iterative cluster (SLIC)~\cite{achanta2012slic}) in the labeling assignment process. We use a small number of superpixels instead of a large number of pixels to solve the labeling problem by SLIC. The motion model with each label is associated to the center of each superpixel in order to assign a label to all the pixels inside this superpixel. SLIC~\cite{achanta2012slic} takes into account the position and color of the pixels, without combining depth information which is essential because of the high discrimination in the three-dimensional space. In this paper, we consider the position, color and depth of each superpixel and average those of the pixels inside it. The distance metric for clustering of the superpixel is given by 

\begin{equation}\label{eq6} 
D=\frac{\sum_{n=1}^3{(u_i(n)-u_c(n))^2}}{N_u^2}+\frac{{(x_i-x_c)^2}+{(y_i-y_c)^2}}{N_s^2},
\end{equation}
\begin{equation}\label{eq7} 
D^{'}=D+\frac{(\frac{1}{d_i}-\frac{1}{d_c})^2}{N_d^2},
\end{equation}

where $D$ and $D^{'}$ are the distances without and with depth information, respectively. i and c are one pixel and one candidate cluster center, respectively. $u(n)$, [x, y] and d are the values of the color channel, location and depth of the pixel, respectively. ${N_u^2}, N_s^2, N_d^2$ are used to normalize the color, distance and depth proximity, respectively, before the summation. We use a dense stereo matching method (libelas~\cite{geiger2010efficient}) to get the depth information of the pixels in the stereo image. Combing the depth information can prevent the masks from growing outside of the object bounds to some extent. 

Following the VO step, we classify the matched feature points to multiple motion models representing different motions. For each superpixel block $s_i$, $\forall i \in [1,S]$, we count the number of feature points that belong to each motion label inside $s_i$. We then simply take the maximum number of feature points belonging to each motion label inside $s_i$ and associate each superpixel with the motion of a single motion model. If the superpixel block $\overline{s}_i$ does not contain any feature point of a motion model, a K-nearest voting method is used for the label assignment. We characterize each superpixel $s_i$ with the 6D feature vector $f_i$ that encodes the RGB color, 2D location and depth value of its cluster center. The distance metric of each feature vector corresponding to a single superpixel is calculated according to (\ref{eq7}). We count the motion labels of the nearest k superpixel blocks around each superpixel block $\overline{s}_i$, and the motion label $^1T_{M_tM_1}$ with the largest number of k labels is assigned to this superpixel block $\overline{s}_i$ without feature points inside it.

\subsection{Projected mask}
\begin{figure}[t]
	\centering
	\subfigure[] {
		\label{fig:fig3-a}     
		\includegraphics[scale=0.25]{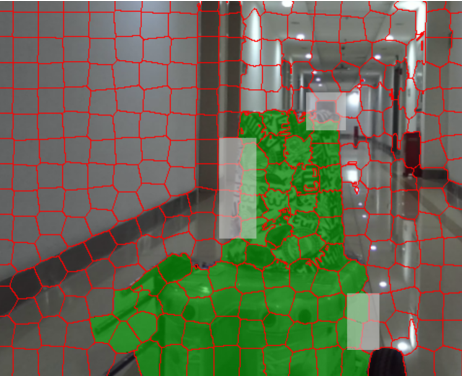}
	} 
	\subfigure[] {
		\label{fig:fig3-b}     
		\includegraphics[scale=0.25]{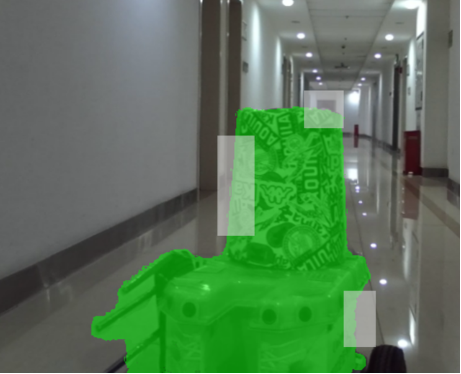}  
	} 
	\subfigure[]{
		\label{fig:fig3-c}     
		\includegraphics[scale=0.17]{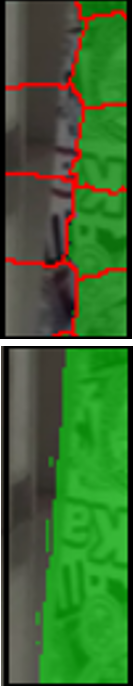} 
	}
	\subfigure[]{
	\label{fig:fig3-d}     
	\includegraphics[scale=0.17]{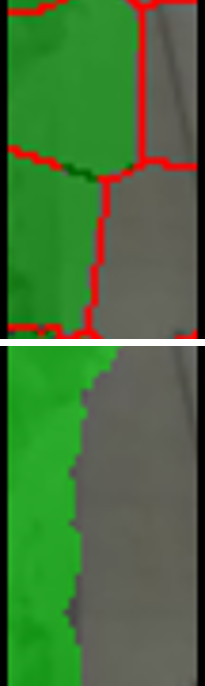} 
    }
	\subfigure[]{
	\label{fig:fig3-e}     
	\includegraphics[scale=0.17]{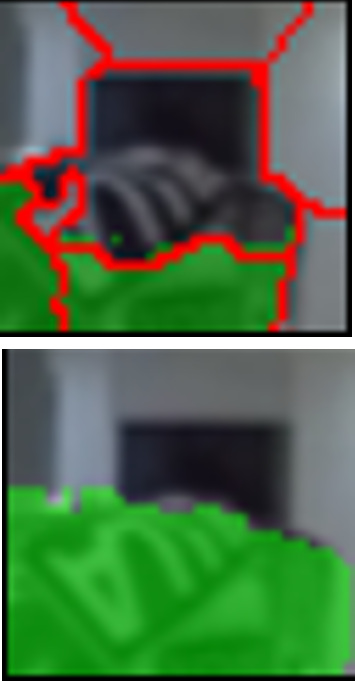} 
	} 
	\caption{(a) is the motion segmentation mask, (b) is the final segmentation mask after fusing the projected mask. The transparent rectangles in (a) and (b) indicates the segment of interest, and they are enlarged in the (c), (d) and (e). (c), (d) and (e) show the comparison of the boundaries of the two segmentation results. While the upper figure shows the motion segmentation mask, and the lower figure shows the final segmentation mask. The results in (c) and (e) improve the effect of insufficient segmentation, and the result in (d) improves the effect of excessive segmentation}
	\label{fig:fig3}
\end{figure}

We complete motion segmentation at the pixel-level and abtain the motion segmentation masks of moving objects at each frame. To compensate for inexact segmentation at the boundary, we utilize the projected masks obtained from transformation from 3D model to 2D image. For each moving object, we project its updated latest 3D model into the 2D image of the current frame using its motion estimation. For each superpixel block ${s^M}_i$ ($\forall i \in [1,S^M]$) assigned to this object's motion label, we calculate the overlap of ${s^M}_i$ relative to the projected mask. Once if this overlap is less than a threshold, ${s^M}_i$ is no longer assigned to this object's motion model, but is assigned to the camera motion model. In our experiments the threshold is $90\%\cdot|{s^M}_i|$, where $|{s^M}_i|$ denotes the number of pixels belonging to the superpixel block ${s^M}_i$. Therefore, as the integrity of the 3D model reconstruction of each object improves, the effect of final segmentation will become better and better theoretically. The motion segmentation mask and the final segmentation mask after fusing the projected mask are shown in the Fig.~\ref{fig:fig3}. Comparison of the boundaries of the two segmentation results is also shown in the Fig.~\ref{fig:fig3} It can be seen that fusing the projected mask improves the effect of inexact segmentation of the boundary well.

\section{Dense Reconstruction}
\label{sec:dense reconstruction}
\begin{figure}[t]
	\begin{center}
		\includegraphics[width=1.0\linewidth]{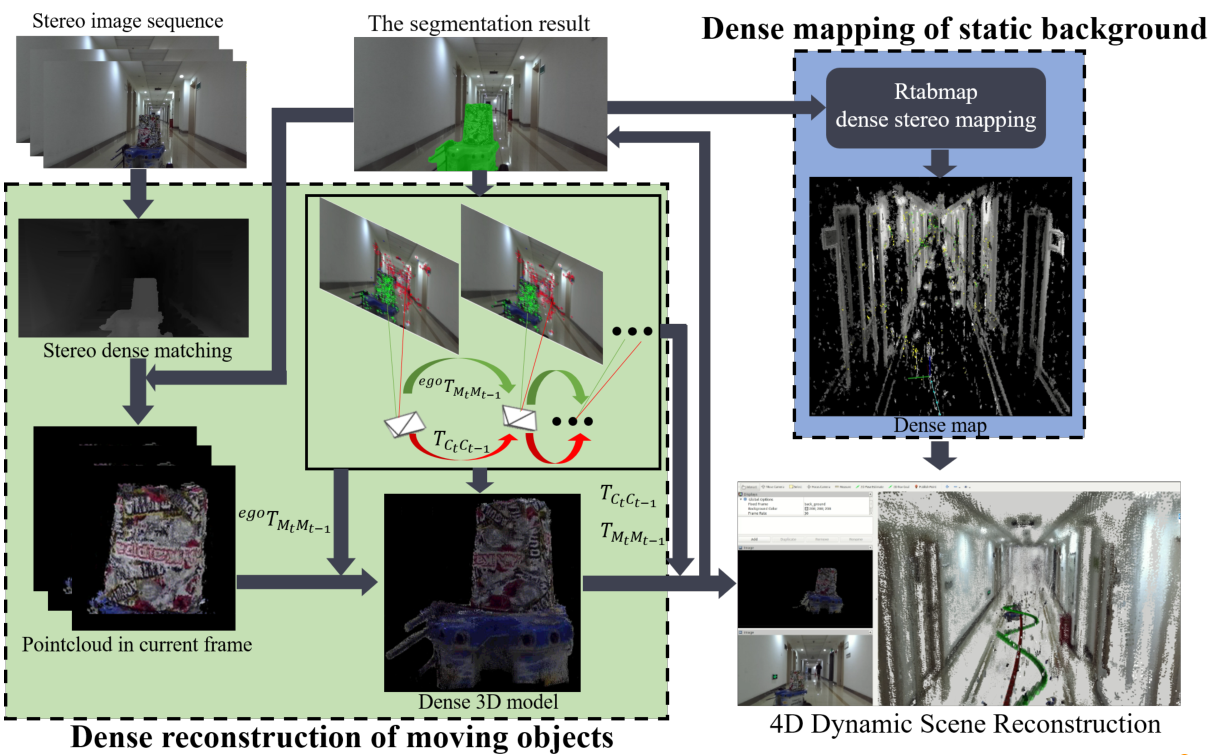}
	\end{center}
	\caption{An overview of the dense reconstruction process. Including dense mapping of the static background and dense reconstruction of the moving objects. A detailed description can be found in Section~\ref{sec:dense reconstruction}.}
	\label{fig:fig2}
\end{figure}

Following the above steps, we estimate the motion trajectories of the camera and the different rigid moving objects, and obtain the accurate masks of moving objects. In this section, we describe how we build a dense 3D point-cloud map of the static background while simultaneously reconstructing 3D models of the moving objects belonging to different motions and improve them over time by merging the newly available 3D point cloud with the existing models. An overview of the dense reconstruction process is shown in Fig.~\ref{fig:fig2}.

\subsection{Dense Mapping of the Static Background}
In the static background mapping, we use only the pixels associated with the motion of the camera and consider all the pixels of moving objects as outliers. This stereo dense mapping strategy is implemented based on a rtabmap~\cite{labbe2019rtab} method. We uses points belonging to the static background to build a dense 3D point cloud map based on the motion trajectory of the camera and the segmentation of moving object masks.

\subsection{Dense Reconstruction of Moving Objects}
Based on the results of the multi-motion segmentation and the moving object mask, we reconstruct a 3D model of each moving object which is associated with a single rigid transformation label. For a moving object whose label is represented by the rigid transformation $T_{M_tM_1}$ in the global reference frame, we first perform stereo dense matching (libelas~\cite{geiger2010efficient}) in the pixel area belonging to this moving object in the RGB stereo image pairs. Its dense point cloud $P_{M_t}$ is obtained based on the known camera extrinsics and intrinsics, and the depth information calculated from the matching in the current frame $t$. Then, in consecutive frames, the point cloud for each frame is stitched using rigid transformation $^{ego}T_{M_tM_{t-1}}$ in two adjacent frames in the egocentric reference frame. $^{ego}T_{M_tM_{t-1}}$ is the motion in the egocentric frame of the moving object and calculated based on the inliers of its motion model at the feature level, instead of using all the points belonging to this object, and which could be more robust and efficient. During the movement, the 3D model of the object is updated by stitching the new point cloud and is transformed into the global reference frame with a rigid transformation $T_{M_tM_1}$. The update of the 3D model can continuously improve not only the segmentation of moving object masks but also the result of its gravity center transformation $T_{init}$ used in the estimation of the motion of moving objects. 

\section{Experiments}
\begin{figure}[t]
	\centering
	\subfigure[] {
		\label{fig:fig4-a}     
		\includegraphics[width=0.42\linewidth]{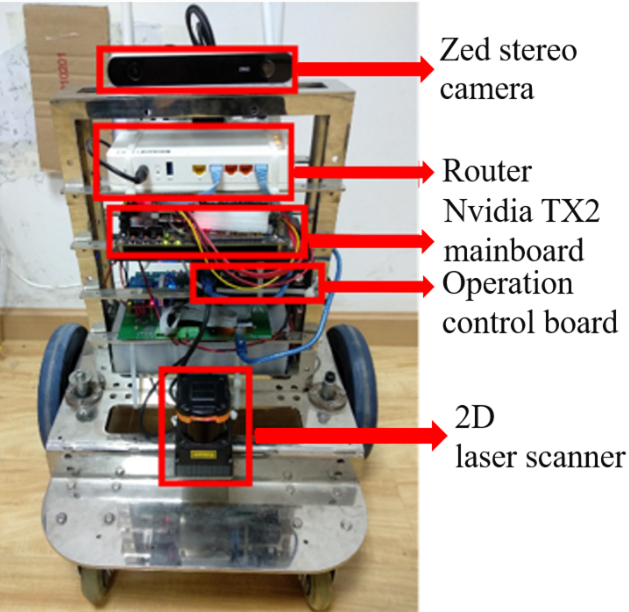}  
	} 
	\subfigure[] {
		\label{fig:fig4-b}     
		\includegraphics[width=0.22\linewidth]{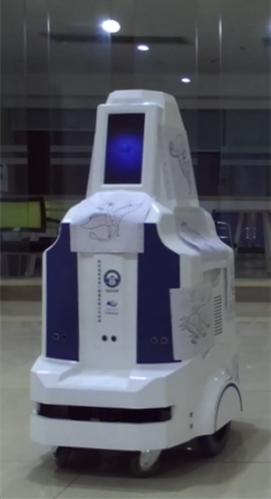}  
	} 
	\subfigure[]{
		\label{fig:fig4-c}     
		\includegraphics[width=0.22\linewidth]{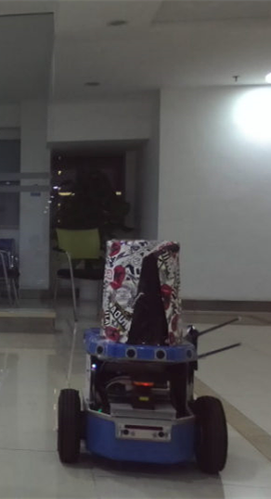} 
	}
	\caption{Mobile acquisition platform for acquiring real dataset in (a). Two moving objects without semantic information in (b) and (c), blue object in (c) uses lidar SLAM to maintain ground-truth data for the trajectory.}
	\label{fig4}
\end{figure}

We carried out a quantitative evaluation on real sequences with ground truth data. As shown in the Fig.~\ref{fig:fig4-a}. For the mobile acquisition platform, a ZED stereo camera was used to obtain the real image pair sequences. The moving objects were two mobile platforms without semantic information. A Hokuyo, 2D laser scanner is mounted on both the mobile acquisition platform and the blue mobile platform shown in Fig.~\ref{fig:fig4-c}. The poses obtained using Google's Cartographer~\cite{hess2016real} algorithm were recorded to obtain ground-truth data for the trajectories of both the mobile acquisition platform and the blue mobile platforms. In order to verify the innovative nature of the method proposed in this paper, several experimental scenarios were carried out. Because the methods mentioned in the related work section that also deal with moving objects are all based on using RGB-D sensors except MVO method~\cite{judd2018multimotion}, we did not compare them to our stereo method. The source code of the only stereo method-MVO is not publicly available, at the same time it only completed part of our task. So we cannot compared MVO against our method.

\begin{figure}[t]
	\centering
	\subfigure[] {
		\label{fig:fig5-a}     
		\includegraphics[width=0.285\linewidth]{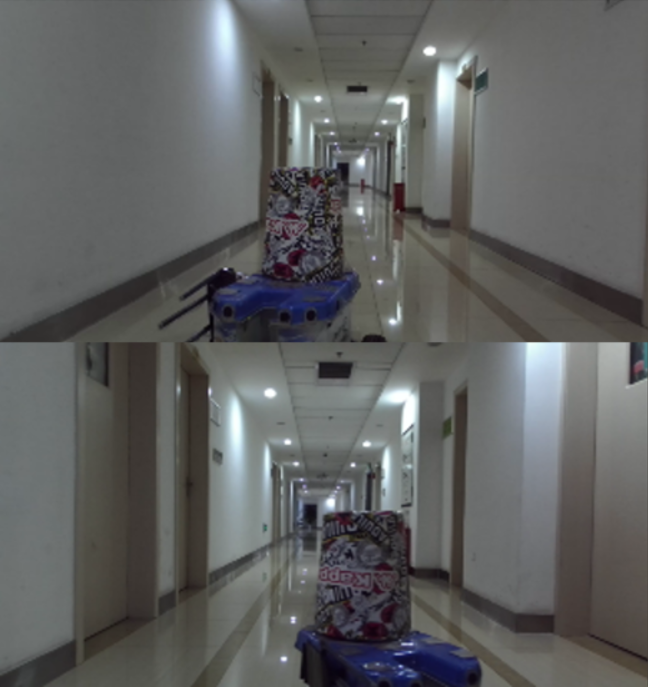}  
	} 
	\hspace{-0.45cm}
	\subfigure[] {
		\label{fig:fig5-b}     
		\includegraphics[width=0.345\linewidth]{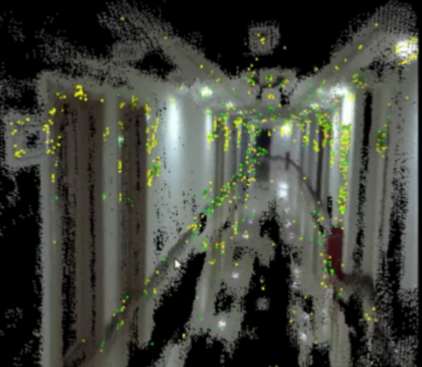}  
	} 
	\hspace{-0.45cm}
	\subfigure[]{
		\label{fig:fig5-c}     
		\includegraphics[width=0.3\linewidth]{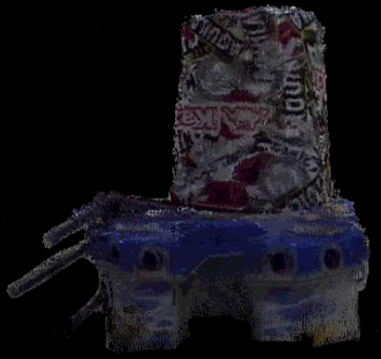} 
	}
	\caption{Both dynamic object and mobile acquisition platform are under wide range movement in indoor corridors of longer than 25m. (b) shows a dense mapping of static background. There are quite a few gaps on the ground in the dense map because this part was always blocked by the moving object and did not appear in the camera's field of view. Dense reconstruction of the moving object is shown in (c).}
	\label{fig:fig5}
\end{figure}

\subsection{4D Dynamic Scene Reconstruction}
This experiment was carried out in indoor corridors of longer than 25m, with both the moving object and the mobile acquisition platform under complex wide range movement (see Fig.~\ref{fig:fig5-a}). The dynamic object and the mobile acquisition platform moved forward along the corridor in tandem. The dynamic object not only moved forward but also made many turns, which increased the difficulty of motion estimation and dense reconstruction. Fig.~\ref{fig:fig5-b} shows the dense mapping of the static background. It can be seen that the proposed method reconstructs the static dense map well in the dynamic corridor environment in the presence of a salient moving object. The result of the dense reconstruction of the moving object is shown in Fig.~\ref{fig:fig5-c}. Since the dynamic object always moved forward in front of the camera, the front part of the object did not appear in the camera's field of view. Therefore, the part that the camera could see behind the dynamic object was reconstructed and the front part was not. Meanwhile, the wheel part of the object is not reconstructed well because the motion model of the wheel part was unique and it was difficult to segment together with the overall dynamic object as a motion model. 

\begin{table}[h]
	\caption{RMSE in position and rotation.} 
	\label{table1}
	\begin{center}
		\begin{tabular}{|l|c|c|}
			\hline
			& Camera & Moving object \\
			\hline\hline
			Position[cm] & 5.14 & 10.81 \\
			Rotation[$^\circ$] & 1.2435 & 2.0472 \\
			\hline
		\end{tabular}
	\end{center}
\end{table}

\begin{figure*}
	\centering
	\subfigure[] {
		\label{fig:fig6-a}     
		\includegraphics[width=0.285\linewidth]{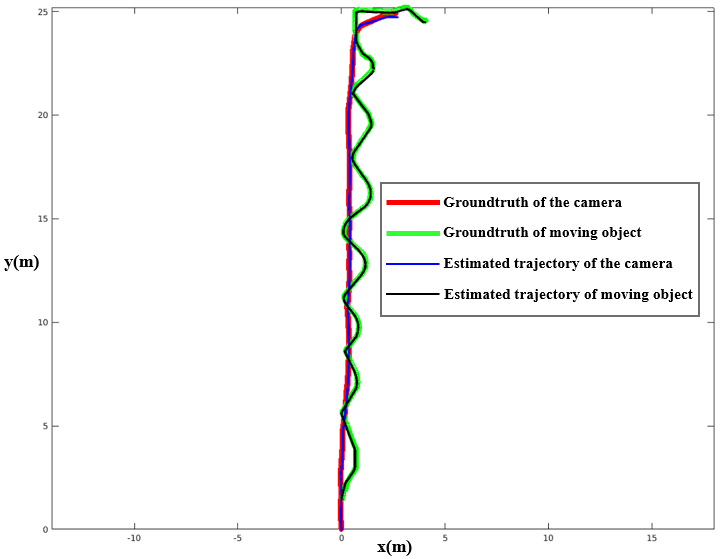}  
	} 
	\hspace{-0.45cm}
	\subfigure[] {
		\label{fig:fig6-b}     
		\includegraphics[width=0.4\linewidth]{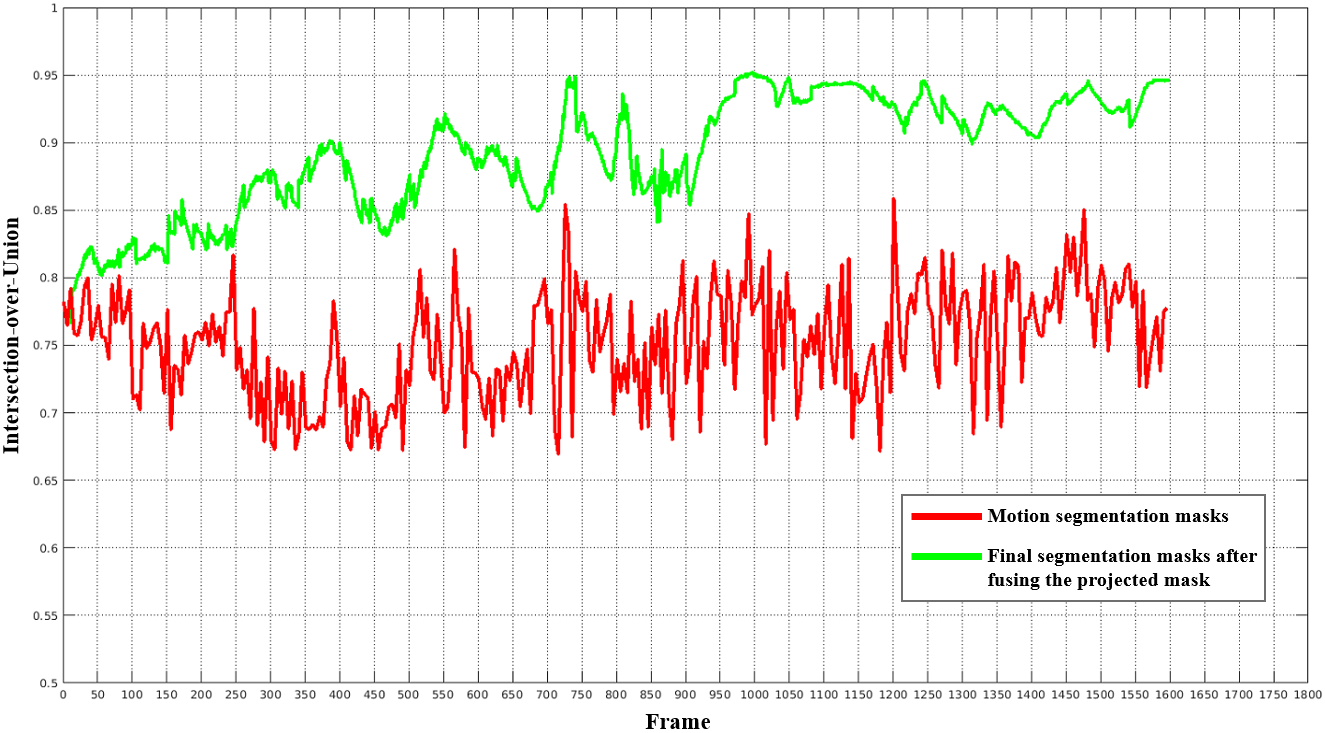}  
	} 
	\hspace{-0.45cm}
	\subfigure[]{
		\label{fig:fig6-c}     
		\includegraphics[width=0.29\linewidth]{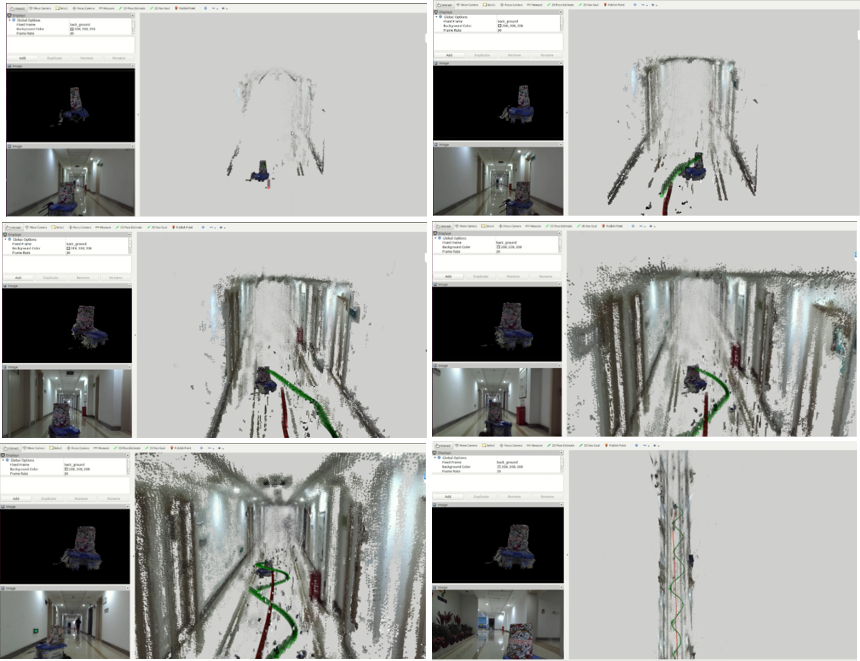} 
	}
	\caption{Comparison between the ground truth and estimated trajectories for the camera and the moving object in (a). The IoU comparison between segmentation masks obtained based on motion segmentation and final segmentation masks after fusing the projected mask in (b). Overall visual display is shown in (c). The red trajectory and the green trajectory are the motion trajectory of the mobile acquisition platform and the moving object, respectively. The point cloud in front of the green cylinders is the result of the current dense reconstruction of the moving object.}
	\label{fig:fig6}
\end{figure*}

\textbf{Trajectory estimation} We compared the estimated and ground-truth trajectories by RMSE (Root Mean Square Error) for the camera and the moving object in a dynamic scene. Results are shown in table~\ref{table1} and Fig.~\ref{fig:fig6-a}. This experiment proved that our method can obtain a good effect of trajectory estimation in the case of both the camera and the moving object undergoing a large range of complex motion.

\textbf{Segmentation} To evaluate the segmentation we compute the commonly used Intersection over Union (IoU) metric for each frame. We acquired a 1800 frame long sequence and provided ground truth 2D annotations manually for the masks of the moving object. Our results are presented in Fig.~\ref{fig:fig6-b}. This figure shows that fusing the projected mask results in more accurate segmentations.


\textbf{Overall visual display} In order to demonstrate the effectiveness of the proposed method, an overall visual display of the proposed method was built based on the ROS platform (see Fig.~\ref{fig:fig6-c}). The picture in the top left of Fig.~\ref{fig:fig6-c} shows that the mobile acquisition platform and the dynamic object are starting to move at the same time. Considering the efficiency of the visual display, the static background point cloud is selectively displayed, so that the visual effect of the static background point cloud after being selectively displayed is worse than the result in Fig.~\ref{fig:fig5-b}. It can be seen that the proposed method estimates the motion trajectories of the camera and the moving object simultaneously, while also reconstructing a dense 3D point cloud of the static background and the moving object. The motion trajectory and the dense 3D model of the moving object are updated continuously by the proposed method while moving. The display is available in \url{https://youtu.be/xw_XFTEZQT0}.


\subsection{Segmentation and Reconstruction of Multiple Objects}
\begin{figure}[t]
	\centering
	\subfigure[] {
		\label{fig:fig7-a}     
		\includegraphics[width=0.73\linewidth]{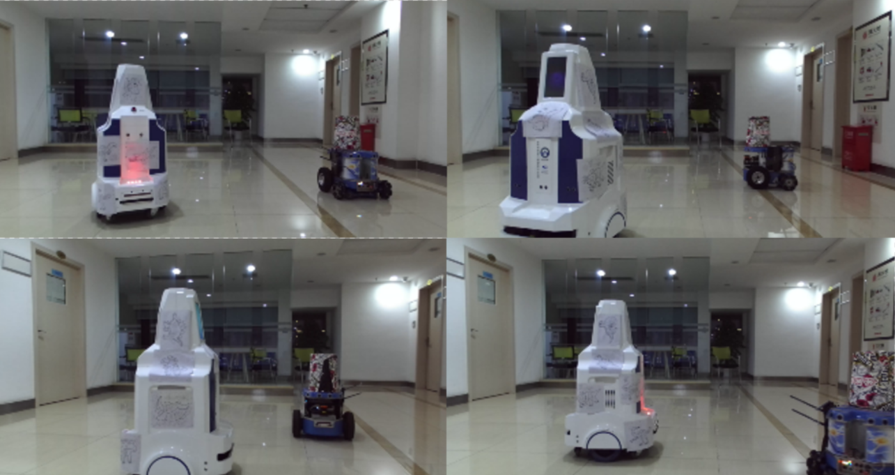}  
	} 
	\hspace{-0.45cm}
	\subfigure[]{
		\label{fig:fig7-b}     
		\includegraphics[width=0.22\linewidth]{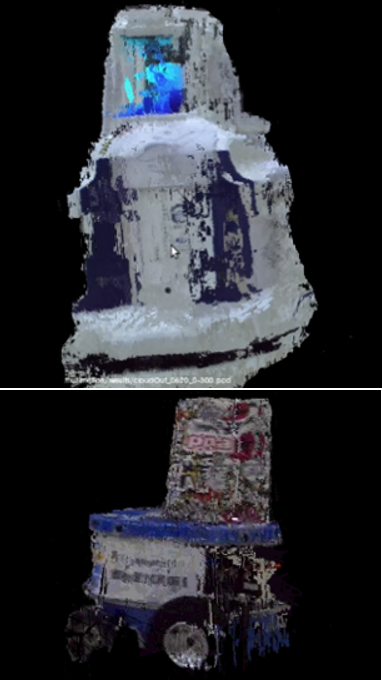} 
	}
	\caption{In (a), the white object on the left rotates in place and the blue object on the right moves along a rectangular path while the mobile platform of acquiring data is also under small movement. The results on dense reconstruction of two moving objects in different motions are shown in (b).}
	\label{fig:fig7}
\end{figure}

This experiment was designed to prove that the proposed method works well in the case of multiple moving rigid objects. In the experimental scene, there were two dynamic objects at any one time (see Fig.~\ref{fig:fig7-a}). The white object on the left rotated in place and the blue object on the right moved along a rectangular path, while the mobile platform of acquiring the data also made under small movements. The results of the dense reconstruction of two moving objects are shown in Fig.~\ref{fig:fig7-b}. The proposed method achieves dense reconstruction of both moving objects, and the texture of the modeled objects is clearly visible. Our method works well on moving objects that not only rotate at a large angle but also translate in a wide range. Multi-motion segmentation and moving objects masks segmentation as well as the IoU measure and the dense reconstruction of two moving objects are available in \url{https://youtu.be/3D6RUd0n-vs}.

\section{Conclusion}
In this paper, we have introduced DymSLAM system, which is a dynamic stereo visual SLAM system that estimates the trajectories of camera and each moving rigid object in the global reference frame while simultaneously reconstructing the dense point-cloud of the moving objects and the static background, in order to reconstruct a 4D point cloud of the dynamic scene. The proposed system can segment motion models of different moving objects by the multi-model fitting approach without semantic cues and obtain the accurate masks after fusing the masks projected from 3D model. Compared to other methods based on semantic segmentation and the use of RGB-D sensors, the proposed method works well even when both the camera and unknown moving objects are undergoing wide range movement -- in our experiment is a corridor of longer than 25m. The resulting system could enable a robot to obtain better scene perception of the environment, allowing it to be employed for high-level tasks, such as obstacle avoidance for dynamic objects. 

In our future work, we will actively explore the mutual benefits of continuous time motion estimation and dense reconstruction, as well the implementation of a pipeline that could be used in real time. Meanwhile, we will try to apply the system to robot autonomous navigation to improve the ability of robots to avoid moving obstacles in dynamic scenes.

{\small
	\bibliographystyle{IEEEtran}
	\bibliography{egbib}
}

\end{document}